\documentclass[letterpaper, 10 pt, journal, twoside]{IEEEtran}

\ifCLASSINFOpdf
  \usepackage[pdftex]{graphicx}

\else

\fi

\usepackage{amsmath}
\usepackage{amsfonts}
\usepackage{booktabs}
\usepackage{tabularx}
\usepackage{multirow}
\usepackage[table]{xcolor}
\usepackage{colortbl}
\usepackage{makecell}
\usepackage{booktabs}
\usepackage{multirow} 
\usepackage{cite}
\usepackage{balance}

\begin{document}

\title{\LARGE \bf
Autonomous FPV Flight with Translational Optical Flow and Uncertainty Mask
}

\author{Yang Deng$^*$, Yu Hu$^*$, Feng Yu, Linzuo Zhang, Danping Zou$^{\dag}$
\thanks{Manuscript received: February 24, 2026; Revised: April 12, 2026; Accepted: June 6, 2026.}
\thanks{This paper was recommended for publication by Editor G. Loianno upon evaluation of the Associate Editor and Reviewers' comments. This work was supported by National Key R\&D Program of China (2022YFB3903802) and National Science Foundation of China (62073214).
} 
\thanks{The authors are with the Shanghai Jiao Tong University, Shanghai 200240, China (e-mail: dpzou@sjtu.edu.cn).}
\thanks{$^*$ These authors contributed equally to this work. }
\thanks{$^{\dag}$ Corresponding author.}
\thanks{Digital Object Identifier (DOI): see top of this page.}
}

\markboth{IEEE Robotics and Automation Letters. Preprint Version. June, 2026}
{Deng \MakeLowercase{\textit{et al.}}: Autonomous FPV Flight with Translational Optical Flow and Uncertainty Mask} 

\maketitle

\begin{abstract}

Autonomous FPV quadrotor flight in complex environments using a monocular RGB camera as the sole exteroceptive sensor remains a fundamental challenge. Recent research has shown that using optical flow as the input of a neural network can achieve end-to-end autonomous flight in cluttered scenes. However, extracting the most relevant information from the flow estimation is the key bottleneck limiting agility and robustness. 
Existing methods struggle to disentangle obstacle-induced optical flow from the ego-motion background flow and suffer from low signal-to-noise ratios near the focus of expansion (FoE).
To address these issues, we decompose the optical flow into translational and rotational components and utilize only the translational flow, which captures scene geometry and depth cues. In addition, we introduce an uncertainty mask derived from inconsistencies between forward and backward flow estimates. This mask highlights obstacle structures, including those within the FoE region. 
Both cues are fed to a control policy trained in a differentiable simulation framework, which enables efficient first-order optimization across perception and control. We validate our approach through extensive experiments in both simulated and real-world forest environments. 
The proposed system achieves robust flight at speeds of up to 13.91 m/s in simulation and 11.79 m/s in real-world tests, with a 93.3\% success rate over 30 real-world trials, nearly doubling the previously reported 6 m/s real-world speed of the monocular-RGB optical-flow UAV obstacle avoidance system.
\end{abstract}
\section{Introduction}


Nature demonstrates the potential of monocular vision for agile flight. Human first-person-view (FPV) pilots can execute aerobatic maneuvers in cluttered environments using monocular RGB images~\cite{kaufmann2023champion}. Likewise, insects such as fruit flies and bees achieve visual localization~\cite{srinivasan1996honeybee}, gap estimation~\cite{ravi2019gap}, tunnel navigation~\cite{baird2012visual}, and obstacle avoidance~\cite{ravi2022bumblebees} with monocular vision despite simple neural systems. These observations suggest that agile flight in cluttered environments may be achievable from monocular images with compact neural policies.

\begin{figure}[thbp]
  \centering
  \includegraphics[width=0.45\textwidth]{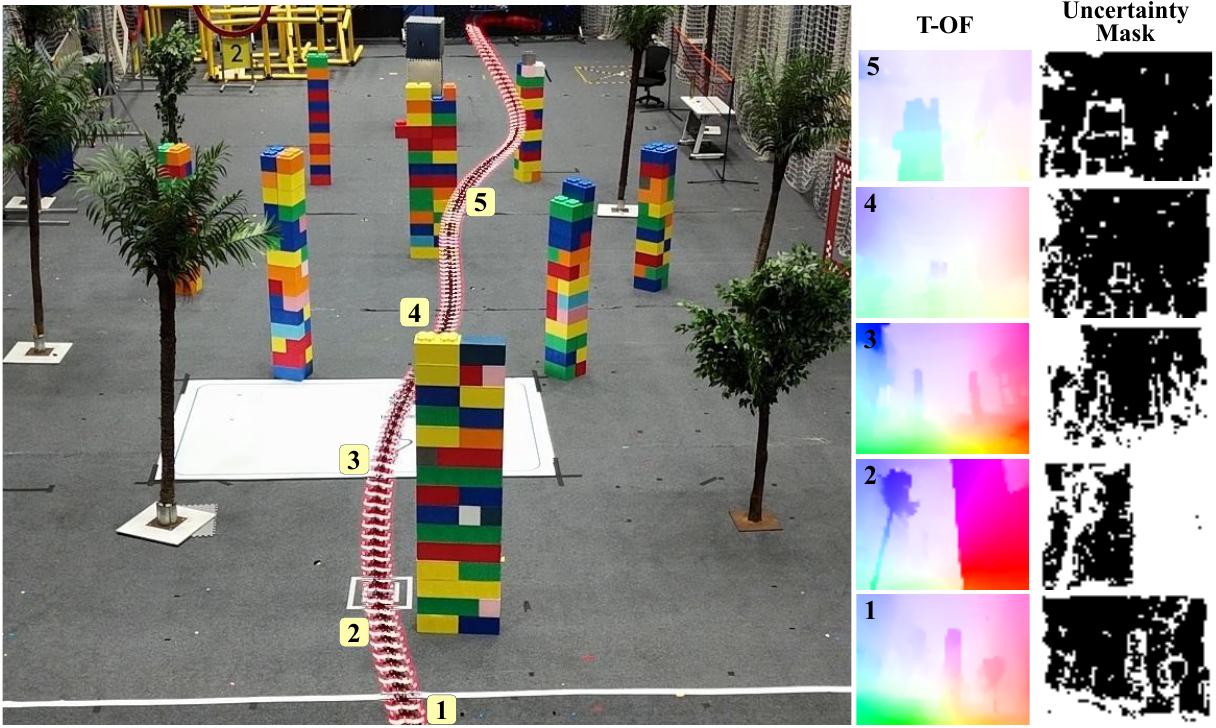}
      \caption{\textbf{Autonomous navigation in cluttered real-world environments using translational optical flow and an uncertainty mask from a single camera:}
      Left: real-world test environment and resulting flight trajectory.
      Middle: representative real-world translational optical flow (T-OF) estimated by GMFlow, shown with HSV-style color encoding, where hue indicates direction and saturation indicates magnitude.
      Right: corresponding real-world uncertainty masks computed via forward--backward optical-flow consistency, with white indicating uncertain area and black indicating certain area.
      Examples on the right correspond to numbered viewpoints (1--5) along the flight direction of the trajectory shown on the left.}
  \vspace{-1.7em}
  \label{fig:cover}
\end{figure}

Prior work has studied monocular obstacle avoidance in FPV settings. Most methods~\cite{Gandhi2017Crashing, sadeghi2017cad2, Loquercio2018DroNet, low2025sous, huang2025flying} use CNN-based visual encoders followed by MLPs that map RGB inputs to control actions. However, they often suffer from suboptimal convergence due to limited data, scene overfitting, or simplified control strategies, and are therefore typically limited to sparse or structured environments at speeds below 3~m/s.

Optical flow, which encodes ego-motion relative to the environment as a 2D vector field, has been applied to UAV takeoff~\cite{Volker2015Nonlinearegomotion}, landing~\cite{cheng2019motion, Croon2021enhancing, HO2024INDI, Herisse2008Hoveringflightandverticallanding}, hovering~\cite{Herisse2008Hoveringflightandverticallanding, HO2024INDI}, and low-speed obstacle avoidance in simple environments~\cite{nelson1989obstacle, serres2017optic, Croonattctrl}. Because these tasks typically involve specific, highly structured forms of optical flow, they can often be handled by relatively simple or task-specific controllers, including PID~\cite{Croon2021enhancing, cheng2019motion, Volker2015Nonlinearegomotion}, MPC~\cite{banday2025event}, and INDI~\cite{HO2024INDI}.

\begin{figure*}[thbp]
  \centering
  \includegraphics[width=0.85\textwidth]{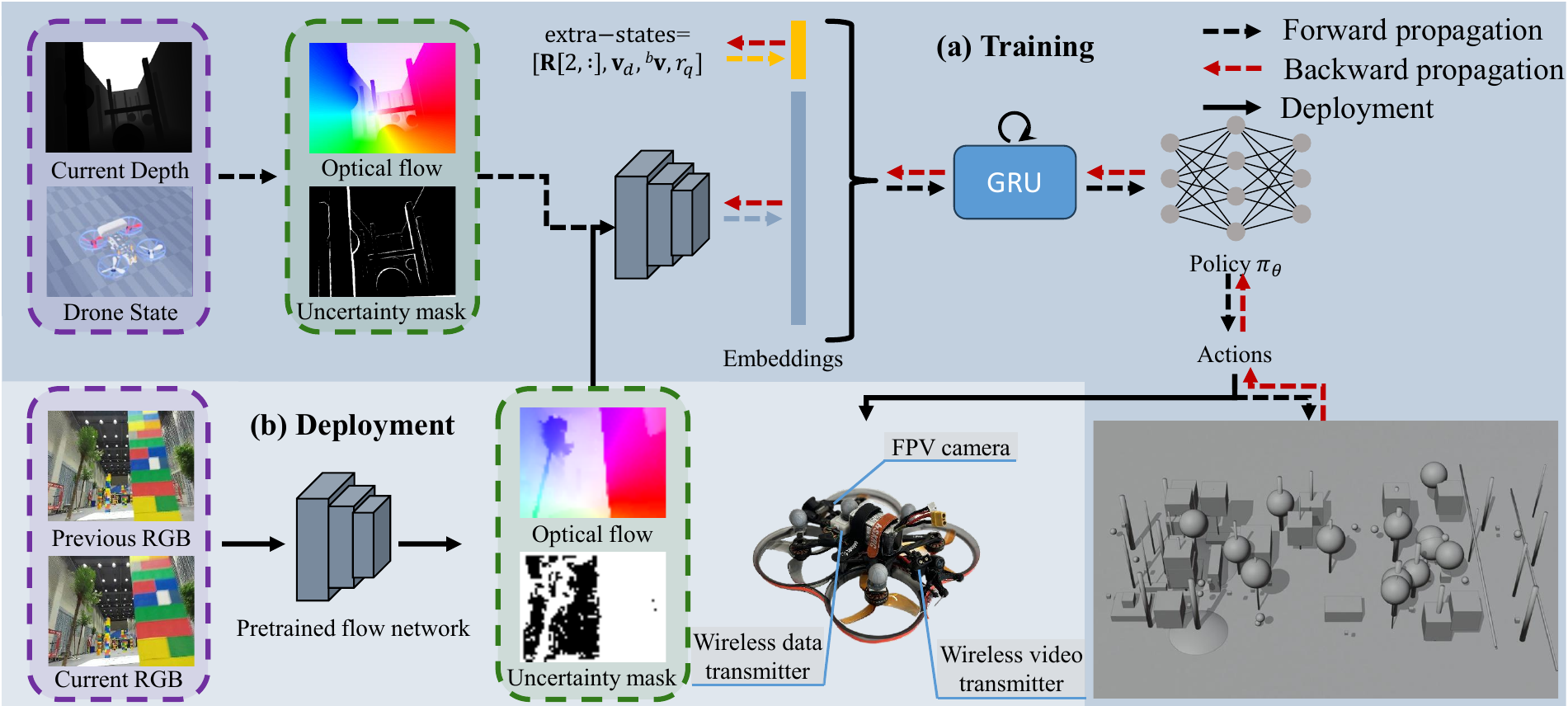}
  \caption{\textbf{Framework of the proposed training and deployment method:} \textbf{(a) Training:} The policy is trained in a CUDA-based differentiable simulator with discrete-time point-mass UAV dynamics. The simulator state is propagated in discrete time(1/15~s), and at each simulated pose, depth is rendered by ray casting while ground-truth optical flow is generated by pixel reprojection between the two camera poses. A CRNN policy takes $(\mathcal{F}_t,\mathcal{M}_t)$ and extra states $[\mathbf{R}[2,:], \mathbf{v}_d, {}^b\mathbf{v}, r_q]$ as input, outputs control commands, and is optimized by backpropagation through the dynamics. \textbf{(b) Deployment:} Two RGB frames are processed by a pretrained optical flow network to estimate $\mathcal{F}_t$ and $\mathcal{M}_t$, which are fed into the trained policy for agile obstacle avoidance in flight.}

  \label{fig:framework}
  \vspace{-1.5em}
\end{figure*}

Beyond low-level flight primitives, optical flow has been explored as a compact representation for learning-based navigation in cluttered environments, improving generalization without explicit metric reconstruction. However, because it comes from image motion rather than direct metric geometry, designing robust control policies from raw 2D flow remains challenging. A recent study~\cite{huyu2025SeingThroughPixelMotion} showed that high-speed obstacle avoidance can be learned directly from raw optical flow using a CRNN trained in a differentiable simulator, demonstrating its potential for agile monocular flight. Nevertheless, applying it to aggressive flight still faces two key limitations: optical flow is highly sensitive to ego-motion, and coupling translational and rotational components can blur the distinction between obstacles and background, especially during rapid rotations; meanwhile, near the Focus of Expansion (FoE), the image point from which translational optical flow radiates outward during forward motion, flow magnitude becomes small, reducing the reliability of obstacle cues in this region.

Motivated by these observations, this work improves the visual representation for monocular RGB obstacle avoidance. Specifically, we use translational optical flow and a flow-derived uncertainty mask as unified policy input: the former reduces rotation-induced ambiguity and provides motion cues more directly related to scene structure, while the latter emphasizes geometrically informative regions, especially near the FoE where raw flow cues are weak. Although each component is classical, their task-driven integration yields a more effective representation for differentiable learning of high-speed monocular obstacle avoidance. The overall training and deployment pipeline is shown in Fig.~\ref{fig:framework}.
In summary, our main contributions are:
\begin{enumerate}
    \item We analyze the limitations of raw optical flow for high-speed monocular RGB obstacle avoidance and identify two primary failure modes: translational--rotational coupling and degraded obstacle observability near the FoE.
    \item We propose a task-oriented visual representation integrating translational optical flow and a flow-derived uncertainty mask as policy input.
    \item We validate the proposed representation in simulation and real-world experiments, demonstrating improved robustness and higher achievable flight speeds in cluttered monocular RGB flight.
\end{enumerate}

\section{Related Work}

\subsection{Monocular Vision for Autonomous UAV Navigation}

Monocular-vision-based UAV navigation has attracted considerable attention because of its low payload, low power consumption, and biological inspiration. Early works typically use CNNs to directly map monocular RGB images to control commands in an end-to-end manner. Representative examples include DroNet~\cite{Loquercio2018DroNet}, which predicts steering angles and collision probabilities, and imitation-learning-based approaches~\cite{Gandhi2017Crashing, sadeghi2017cad2}. Although effective in controlled settings, these methods often suffer from limited generalization, sensitivity to appearance variations, and restricted flight speeds, typically below 3~m/s.


To improve robustness and temporal consistency, recurrent architectures and reinforcement learning (RL) have been explored~\cite{kaufmann2018deep, kaufmann2023champion}. However, learning reliable control policies directly from RGB images remains difficult, since RGB observations are highly redundant, sensitive to appearance variations irrelevant to control, and lack explicit geometric information. Recent works have also used Gaussian scene representations to map RGB renderings from 3D Gaussian splatting to control actions via RL~\cite{low2025sous, huang2025flying}, reducing the RGB-domain sim-to-real gap. Nevertheless, these policies still generalize poorly across environments. These limitations motivate the use of optical flow as a geometry-aware monocular intermediate representation for agile UAV navigation.

\subsection{Optical-Flow-Based Flight Control}

Optical flow provides a compact motion-centric representation and has long been used for UAV flight control, especially in biologically inspired systems. Early studies exploited simple cues such as divergence or expansion for takeoff~\cite{Volker2015Nonlinearegomotion}, hovering, landing~\cite{Herisse2008Hoveringflightandverticallanding}, and low-speed obstacle avoidance~\cite{nelson1989obstacle, serres2017optic}. Although lightweight and interpretable, these methods relied on handcrafted flow measurements and classical controllers, limiting speed and environmental complexity.

Later work improved robustness through advanced frameworks such as MPC~\cite{banday2025event}, INDI~\cite{HO2024INDI}, and observability-aware formulations~\cite{Croonattctrl}. While improving stability and nonlinear handling, these methods assumed reliable optical flow and did not address representation-level limitations. More recent learning-based pipelines further improved real-world robustness, including appearance-aware control~\cite{Croon2021enhancing}, lightweight estimators such as NanoFlowNet~\cite{bouwmeester2023nanoflownet}, and raw-optical-flow-based agile flight in differentiable learning frameworks~\cite{huyu2025SeingThroughPixelMotion}. These results highlight the potential of optical flow for high-speed navigation, but most approaches still operate on raw optical flow and thus inherit translational--rotational coupling and reduced reliability near the focus of expansion (FoE).

These limitations are also consistent with broader computer vision findings. Classical motion-field analysis shows that ego-motion-induced image motion can be decomposed into translational and rotational components~\cite{longuet1980interpretation, fleet2006optical}, while related rotation-compensation ideas have recently been revisited for motion interpretation under strong camera motion~\cite{bideau2024right}. In parallel, forward--backward consistency and related confidence cues are widely used to identify occlusions, unreliable correspondences, and low-confidence regions~\cite{meister2018unflow, SUN2023Decoupled, jeong2024ocai}. 
Motivated by these ideas, we build on flow decomposition and consistency-based confidence cues, which have primarily been used in prior work to improve optical-flow estimation quality. Different from these optical-flow-centered uses, we integrate translational optical flow and a flow-derived uncertainty mask into a differentiable learning framework as task-oriented policy inputs for robust and agile monocular RGB quadrotor flight in cluttered environments.

\section{Methodology}
\subsection{Problem Formulation}
The quadrotor is modeled as a discrete-time dynamical system with continuous state and control spaces \(\mathcal{X}\) and \(\mathcal{U}\). At time step \(t\), the state and control are denoted by \(\mathbf{x}_t \in \mathcal{X}\) and \(\mathbf{u}_t \in \mathcal{U}\), respectively. The observation \(o_t \in \mathcal{O}\) is generated by the observation model \(h\), i.e., \(o_t = h(\mathbf{x}_t, \mathbf{u}_t)\), while the dynamics are governed by the transition function \(f\colon \mathcal{X} \times \mathcal{U} \rightarrow \mathcal{X}\). Following \cite{zhang2025learning}, we adopt a simplified point-mass model with system identification for control and dynamics calibration. In the differentiable simulator, the translational state is \(\mathbf{x}_t=(\mathbf{p}_t,\mathbf{v}_t)\), where \(\mathbf{p}_t,\mathbf{v}_t \in \mathbb{R}^3\); acceleration is computed from the control output, and the attitude matrix is updated separately for attitude-related variables and command representation. In our setting, the task objective is to fly through cluttered environments while avoiding collisions and tracking commanded flight speed.

Our policy \(\pi_\theta\) takes translational optical flow \(\mathcal{F}_t\) and the uncertainty mask \(\mathcal{M}_t\) as inputs and outputs the control command \(\mathbf{u}_t\). The optimal parameters \(\theta^*\) are learned by minimizing the cumulative loss over a horizon of length \(N\):
\begin{align}
    \theta^* &= \arg\min_{\theta} \; \mathcal{L}_{\theta}, \\
    \mathcal{L}_{\theta} &=
      \sum_{t=0}^{N-1} l_t(\mathbf{x}_t, \mathbf{u}_t)
    =
      \sum_{t=0}^{N-1} l_t\bigl(\mathbf{x}_t, \pi_{\theta}(\mathcal{F}_t, \mathcal{M}_t)\bigr),
\end{align}

where \(l_t(\cdot)\) denotes the instantaneous loss at time step \(t\). The parameters are updated by gradient descent:
\begin{align}
    \theta \leftarrow \theta - \gamma \nabla_{\theta}\mathcal{L}_{\theta},
\end{align}
where $\gamma$ is the learning rate. During training, optical flow is rendered by pixel reprojection using rendered depth and relative camera poses. For real-world deployment, we use GMFlow~\cite{xu2022gmflow} to estimate flow from consecutive RGB images. Unless otherwise stated, backward optical flow with the current frame as reference is used throughout.

\begin{figure}
    \centering
    \includegraphics[width=0.42\textwidth]{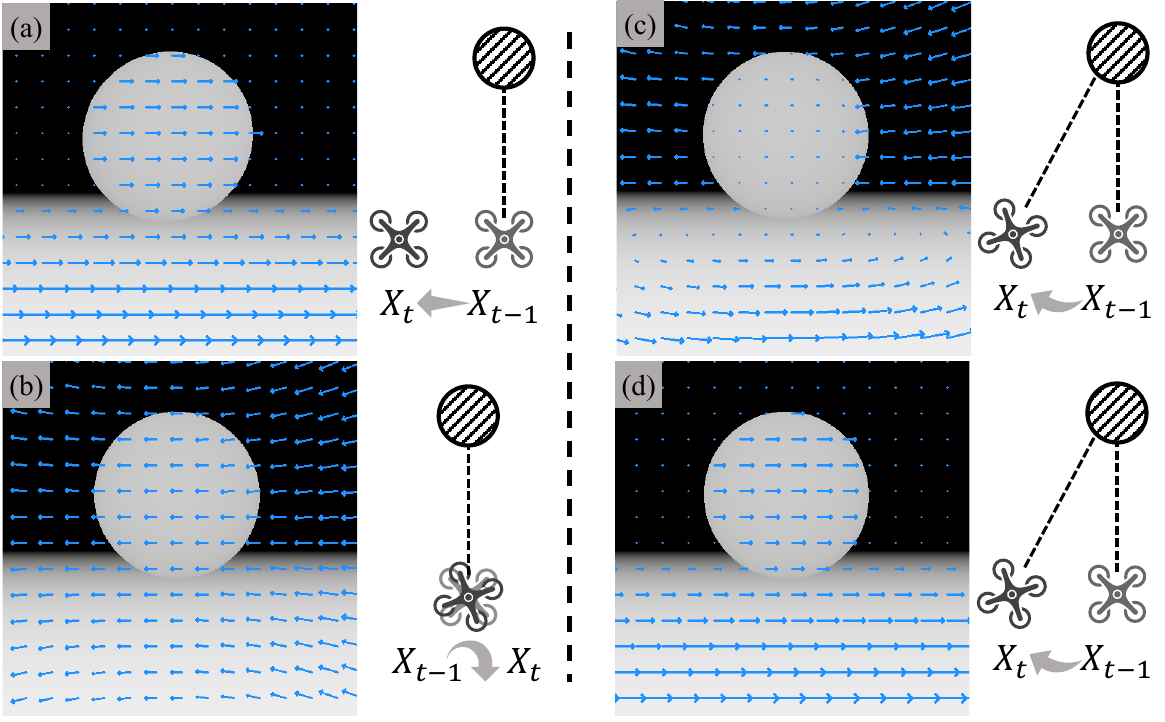}
    \caption{\textbf{Translational vs. Raw Optical Flow under Composite Motion:} Blue arrows indicate optical flow direction and magnitude. (a) Flow under pure leftward translation. (b) Flow under pure clockwise yaw rotation. (c) Raw optical flow under combined leftward translation and clockwise yaw rotation. (d) Translational optical flow extracted from (c) after removing the rotational component, while preserving the same underlying motion.}
    \label{fig:method_tflow}
    \vspace{-1.0em}
\end{figure}

\subsection{Translational Optical Flow as Structural Cues}

We first establish the analytical relationship among scene depth, optical flow, and camera ego-motion.
Under the pinhole camera model, the instantaneous image motion induced by camera translation and rotation is described by the classical motion field equation:
\begin{equation}\label{eq:flow_depth_spd}
    \begin{aligned}
        \begin{bmatrix}
            \dot{\overline{p}}_{x} \\
            \dot{\overline{p}}_{y}
        \end{bmatrix}
    &=\frac{1}{p_{z_b}}
    \begin{bmatrix}
        -1 & 0 & \overline{p}_{x} \\
        0 & -1 & \overline{p}_{y}
    \end{bmatrix}
    {}^{b}\mathbf{v}
    \\
    &\quad+
    \begin{bmatrix}
        \overline{p}_{x}\overline{p}_{y} & -(1+\overline{p}_{x}^{2}) & \overline{p}_{y} \\
        1+\overline{p}_{y}^{2} & -\overline{p}_{x}\overline{p}_{y} & -\overline{p}_{x}
    \end{bmatrix}
    {}^{b}\boldsymbol{\omega}
    \end{aligned}
\end{equation}
where $(\overline{p}_x,\overline{p}_y)=(X/Z,Y/Z)$ denote normalized image coordinates of a 3D point in the camera frame,
$(\dot{\overline{p}}_x,\dot{\overline{p}}_y)$ are the corresponding optical flow components,
$p_{z_b}$ is the depth along the optical axis, and
${}^{b}\mathbf{v}$ and ${}^{b}\boldsymbol{\omega}$ represent the translational and angular velocities of the camera expressed in the body frame.

Equation~\eqref{eq:flow_depth_spd} decomposes optical flow into a depth-dependent translational term and a depth-independent rotational term, exposing the coupling between scene geometry and camera ego-motion. Consistent with prior observations~\cite{huyu2025SeingThroughPixelMotion}, translational flow encodes scene depth and structure, whereas rotational flow mainly reflects camera angular motion. During obstacle-approach maneuvers, however, rotation can make the flow of nearby obstacles resemble that of distant background regions dominated by translation, as shown in Fig.~\ref{fig:method_tflow}(a) and (c). This ambiguity hinders the policy from separating obstacle-induced motion from ego-motion-induced background flow, making structural or depth cues difficult to extract from raw optical flow during agile flight. Inspired by biological vision systems that selectively exploit different flow components for navigation~\cite{egelhaaf2023optic}, we explicitly remove the rotational component, yielding a flow field consistent with pure translational motion across diverse ego-motion patterns and improving obstacle--background separability, as illustrated in Fig.~\ref{fig:method_tflow}(a) and (d).


We compute translational optical flow by subtracting the rotational component from the raw optical flow.
The rotational flow can be approximated using a homography induced by pure camera rotation:
\begin{align}
    \mathbf{H} &= \mathbf{K}\mathbf{R}\mathbf{K}^{-1} \\
    \mathbf{p}' &= \mathbf{H}\mathbf{p} \\
    \boldsymbol{\mathcal{F}}_r(\mathbf{p}) &= (u, v) = \mathbf{p}' - \mathbf{p} \\
    \boldsymbol{\mathcal{F}}_t &= \boldsymbol{\mathcal{F}} - \boldsymbol{\mathcal{F}}_r
\end{align}
where $\mathbf{p}=[x,y,1]^\top$ and $\mathbf{p}'$ are the homogeneous pixel coordinates in the current image $I_{\mathrm{curr}}$
and its corresponding warped location under pure rotation, respectively.
$\mathbf{K}$ denotes the camera intrinsic matrix, $\mathbf{R}\in\mathrm{SO}(3)$ is the relative rotation from $I_{t-1}$ to $I_{t}$, and $\boldsymbol{\mathcal{F}}$, $\boldsymbol{\mathcal{F}}_r$, and $\boldsymbol{\mathcal{F}}_t$ represent the raw, rotational, and translational optical flow fields. 

This homography-based formulation efficiently compensates for the dominant rotation-induced component in raw optical flow and improves obstacle--background separability. It is exact under pure rotation and approximate in general 3D scenes with coupled translation and rotation. In our setting, the objective is not exact geometric decomposition but a control-oriented representation that suppresses rotational contamination. This approximation is effective in practice, although its accuracy may deteriorate in highly non-planar scenes, at large depth discontinuities, or with inaccurate rotation estimates.

For real-world deployment, given two consecutive RGB frames $(I_{t-1}, I_t)$, we estimate the relative rotation $\mathbf{R}$ using SURF~\cite{bay2008speeded} and warp $I_{t-1}$ to the current viewpoint via $\mathbf{H}=\mathbf{K}\mathbf{R}\mathbf{K}^{-1}$. The aligned pair $(\tilde{I}_{t-1}, I_t)$ is then fed to GMFlow~\cite{xu2022gmflow} to estimate translational optical flow $\boldsymbol{\mathcal{F}}_t$, improving correspondence under strong rotations. We adopt SURF mainly because it estimates the relative rotation directly from the same image pair used for flow estimation, thereby preserving consistency in both synchronization and coordinate frame. IMU-based rotation can also be used when camera--IMU synchronization and extrinsic calibration are accurate.

\begin{figure}
    \centering
    \includegraphics[width=0.4\textwidth]{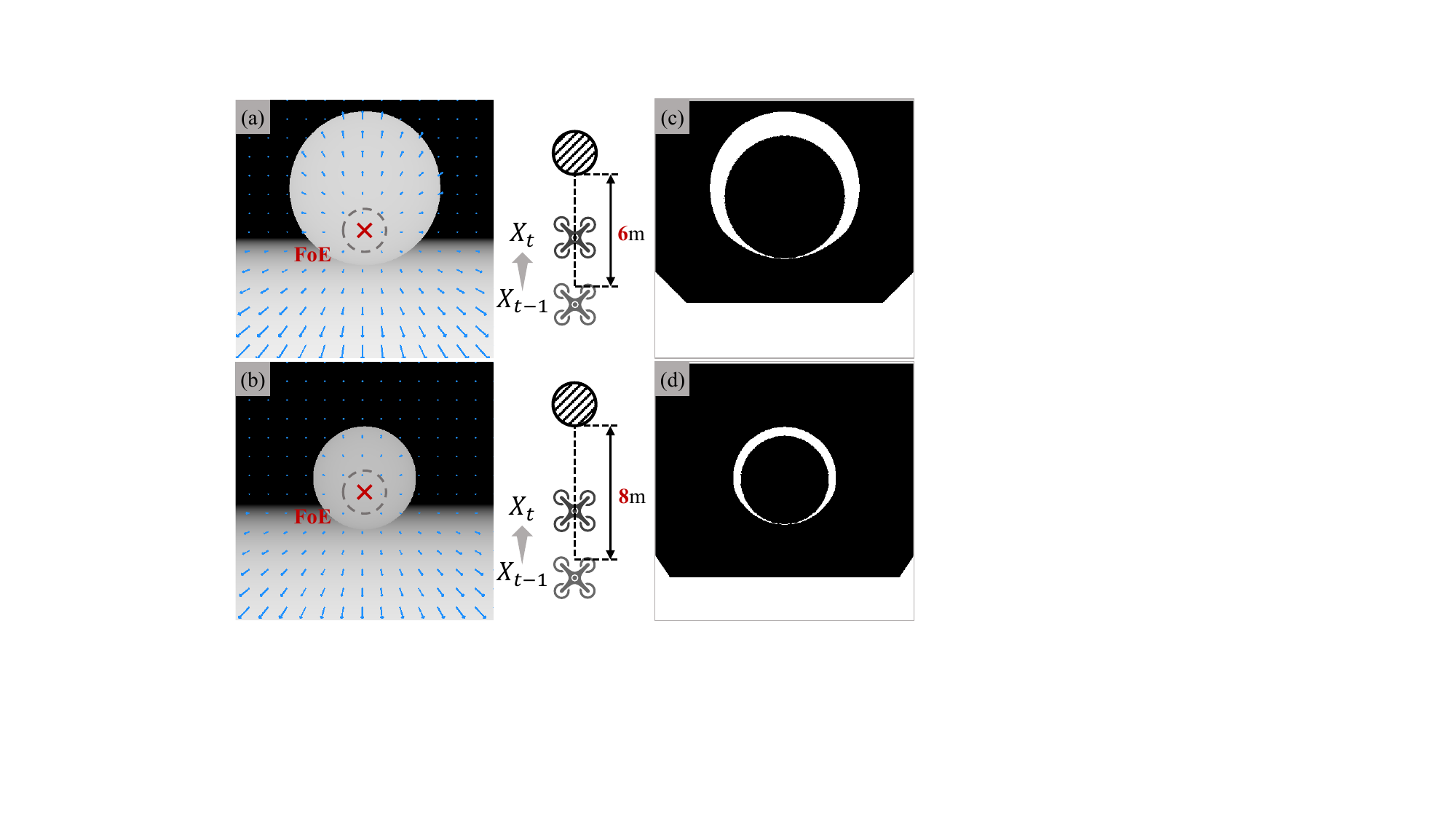}
    \caption{\textbf{Illustration of translational optical flow and its uncertainty mask:} (a) and (b) show translational optical flow when the UAV is 6~m and 8~m from an obstacle, respectively, after moving forward by 1~m. Near the FoE, the flow magnitude is small and hard to distinguish from the background, which may cause collisions. (c) and (d) show the corresponding uncertainty masks, which highlight obstacle contours and provide additional geometric cues, including the ground plane.}
    \label{fig:method_occ}
    \vspace{-1.0em}
\end{figure}

\subsection{Uncertainty Mask as Additional Cues}

Real-world optical flow fields are often corrupted by noise and estimation errors, making structural inference based solely on flow magnitude unreliable, particularly near the FoE. To mitigate this issue, we incorporate a flow uncertainty mask as a complementary geometric cue to improve perception robustness during agile flight. 
Flow uncertainty typically arises from two sources: depth discontinuities at scene boundaries and temporal scene entry or exit events within the camera view, both of which lead to inconsistent flow correspondences. As illustrated in Fig.~\ref{fig:method_occ}, optical flow near the FoE is often small and weakly correlated with obstacle distance. In contrast, the uncertainty mask provides more explicit geometric cues for obstacle localization at both near and far ranges. Moreover, the mask also encodes ground-related geometric information induced by the robot’s own motion, further enriching the structural representation available to the policy network. A real-world example of the uncertainty mask is shown in Fig.~\ref{fig:cover} (right).

The flow uncertainty mask is computed using forward--backward optical flow consistency:
\begin{equation}\label{uncertainty}
\small
\begin{aligned}
    \mathbf{e}(\mathbf{x}) &= \left\| \mathbf{\mathcal{F}}_{f}(\mathbf{x}) + \mathbf{\mathcal{F}}_{b}\left(\mathbf{x} + \mathbf{\mathcal{F}}_{f}(\mathbf{x})\right) \right\|_2 
    \\
    \mathcal{M}(\mathbf{x}) &=
    \begin{cases}
    1, & \text{if } \mathbf{e}(\mathbf{x}) > \alpha\\
    0, & \text{otherwise}
    \end{cases}
\end{aligned}
\end{equation}

where $\boldsymbol{\mathcal{F}}_{f}(\mathbf{x})$ and $\boldsymbol{\mathcal{F}}_{b}(\mathbf{x})$ denote the forward optical flow, and $\alpha$ is a threshold controlling mask sensitivity. $\mathcal{M}(\mathbf{x})=1$ indicates a pixel with large consistency error and thus unreliable or geometrically informative correspondence, while $\mathcal{M}(\mathbf{x})=0$ indicates a more reliable match.

Intuitively, a pixel is first propagated to frame $t{+}1$ using the forward flow and then mapped back using the backward flow. A large discrepancy between the returned position and the original location indicates an unreliable pixel. In training and simulation with ground-truth optical flow, we set $\alpha = 0.01$. For real-world deployment, where optical flow is estimated by a pretrained network and exhibits higher noise levels, a larger threshold $\alpha = 0.1$ is used to avoid overly aggressive masking. We empirically find that system performance remains stable over a reasonable range of $\alpha$.

\subsection{Network Design and Policy Training}
\subsubsection{\textbf{Network architecture}}

We adopt a convolutional recurrent neural network (CRNN) architecture following~\cite{zhang2025learning}. The network comprises a convolutional encoder that extracts hierarchical visual features from translational optical flow and uncertainty mask, followed by a gated recurrent unit (GRU)~\cite{cho2014learning} for temporal fusion. Each frame is processed by four convolutional layers with $(\text{channels}, \text{kernel}, \text{stride})$ of $(32,2,2)\!\rightarrow\!(64,3,2)\!\rightarrow\!(128,3,2)\!\rightarrow\!(192,3,2)$. The resulting feature map is flattened and projected into a 192-dimensional embedding, which is concatenated with UAV proprioceptive states and fed into the GRU and multilayer perceptrons (MLPs) to produce control actions. The policy network contains approximately 1.49M trainable parameters.

\subsubsection{\textbf{Action Space}}
The policy outputs high-level commands at 15~Hz, matching the image update rate. Although the high-level policy runs at 15~Hz, it is trained through trajectory rollout in a differentiable simulator, where future-state losses are backpropagated to earlier actions. This gives the learned policy a short-horizon predictive character rather than a purely myopic reactive behavior. The action space is four-dimensional, $\mathbf{u} = [c, r, p, y]^\top$, where $c$ is the mass-normalized collective thrust, and $r$, $p$, and $y$ are the desired roll, pitch, and yaw angles in the world frame. The world frame is initialized from the UAV state before takeoff and is used only for attitude-related variables and command representation. This thrust-and-attitude formulation fully specifies 3D UAV motion while preserving full control authority. It also matches the dynamics model used in training, enabling stable policy learning and direct deployment on real flight controllers without additional planning layers.

\subsubsection{\textbf{Observation Space}}

The observation at time step $t$ consists of a low-resolution $48 \times 64$ translational optical flow map $F_t$, a max-pooled $48 \times 64$ uncertainty mask $M_t$, and additional state inputs: desired velocity $\mathbf{v}_d$, body-frame velocity ${}^b\mathbf{v}$, quadrotor radius $r_q$, and the third row of the attitude rotation matrix $\mathbf{R}[2,:]$, where $\mathbf{R} \in SO(3)$ denotes the rotation from the body frame to the world frame. The desired velocity $\mathbf{v}_d$ serves as a task-conditioning variable that specifies the commanded flight-speed regime. The policy does not observe the full simulator state, but only the visual inputs and this small set of extra-state variables. To improve adaptability across flight regimes, $\mathbf{v}_d$ is randomly sampled from a predefined range during training, and environments with different obstacle densities are generated accordingly.

\subsubsection{\textbf{Loss Function}}
To achieve agile and safe UAV navigation in cluttered environments, we follow \cite{huyu2025SeingThroughPixelMotion} and optimize the policy with a weighted sum of objective terms:
\begin{equation}
\label{eq:total_loss}
L = \lambda_v \mathcal{L}_v
  + \lambda_c \mathcal{L}_c
  + \lambda_a \mathcal{L}_a
  + \lambda_j \mathcal{L}_j
\end{equation}
where $\lambda_v$, $\lambda_c$, $\lambda_a$, and $\lambda_j$ are the weights for the velocity, collision, acceleration, and jerk losses, respectively. Following \cite{huyu2025SeingThroughPixelMotion}, we adopt the same loss structure and initialize the coefficients accordingly, then empirically tune them to balance velocity tracking, collision avoidance, and motion smoothness in our setting. In our implementation, $\lambda_v = 1.0$, $\lambda_c = 3.0$, $\lambda_a = 0.015$, and $\lambda_j = 0.003$. The additional parameters in the collision loss are fixed to $\beta_1 = 7.0$ and $\beta_2 = -32$.

The individual loss terms are defined as follows:
{
\small
\begin{align}
    \mathcal{L}_v&=\frac{1}{T} \sum_{k=1}^T \text{Smooth L1}
    (\|\mathbf{v}_k^{d}-\bar{\mathbf{v}}_k \|_2,0)\\
    \mathcal{L}_c&=\frac{1}{T} \sum_{k=1}^T \left[v_k^c \max(1-(d_k-r_q),0)^2+\beta_1 \ln(1+e^{\beta_2 (d_k-r_q)})\right]
    \\
    \mathcal{L}_a&=\frac{1}{T} \sum_{k=1}^T \|\mathbf{a}_k \|^2
    \qquad
    \mathcal{L}_j=\frac{1}{T-1} \sum_{k=1}^{T-1} \|\frac{\mathbf{a}_k-\mathbf{a}_{k+1}}{\Delta t}\|^2
\end{align}
}

The velocity loss $\mathcal{L}_v$ penalizes deviations between the desired velocity $\mathbf{v}_k^{d}$ and the executed velocity $\bar{\mathbf{v}}_k$, where $\bar{\mathbf{v}}_k$ is computed by a moving-average filter over instantaneous velocities to suppress high-frequency oscillations and improve training stability.

The collision loss $\mathcal{L}_c$ penalizes insufficient obstacle clearance, where $d_k$ denotes the minimum Euclidean distance to the nearest obstacle at time step $k$, and $r_q$ is the quadrotor radius. The factor $v_k^c$ adaptively increases the proximity penalty at higher speeds, while the logarithmic barrier provides a smooth and differentiable safety margin. 

The acceleration and jerk losses, $\mathcal{L}_a$ and $\mathcal{L}_j$, regularize motion smoothness by penalizing large accelerations and abrupt changes in acceleration, thereby encouraging dynamically feasible, stable, and hardware-friendly trajectories.
\section{Experiments}\label{sec:experiment}

\begin{figure}
    \centering
    \includegraphics[width=0.45\textwidth]{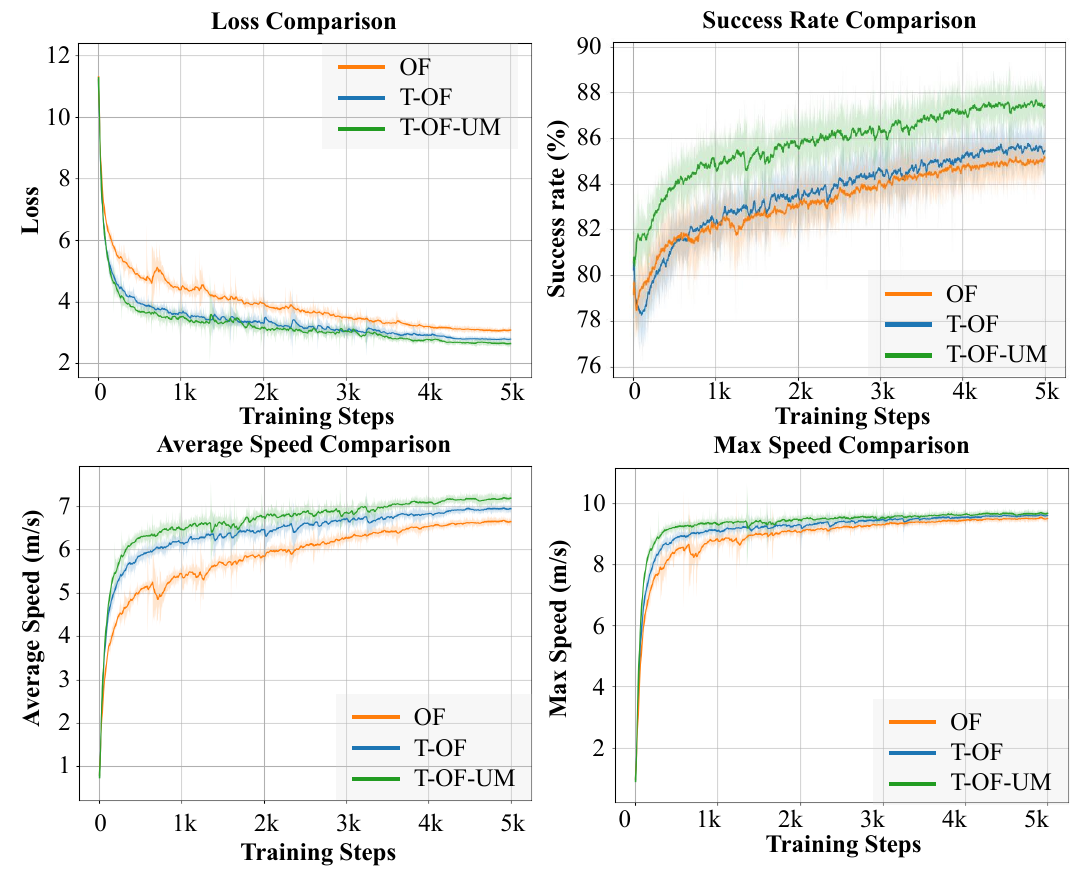}
    \caption{\textbf{Training performance comparison:} Training curves for raw optical flow (OF), translational optical flow (T-OF), and translational optical flow with uncertainty mask (T-OF-UM). Removing the rotational component primarily improves achievable flight speed, while incorporating the uncertainty mask further enhances robustness during agile flight.}
    \label{fig:training}
    \vspace{-0.8em}
\end{figure}

\begin{figure*}
    \centering
    \includegraphics[width=0.94\textwidth]{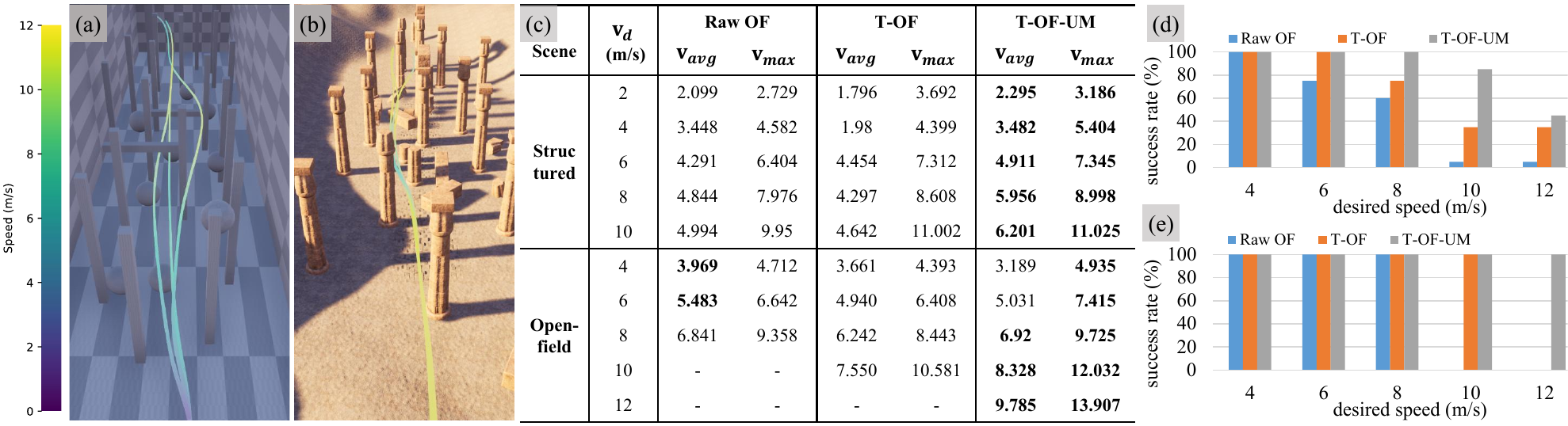}
    \caption{\textbf{AirSim simulation experiments and ablation study:} (a) and (b) show representative trajectories in a structured cluttered environment and a large-scale open-field environment, demonstrating generalization across scene layouts. (c) reports quantitative ablation results in the environments of (a) and (b), where $\mathbf{v}_{avg}$ and $\mathbf{v}_{max}$ denote the average and maximum flight speeds, respectively. (d) and (e) show the corresponding success rates under different desired-speed settings. Policies using translational optical flow outperform raw-flow policies in success rate, and adding the uncertainty mask further improves both success rate and flight speed, consistent with the training results.}
    \label{fig:airsim_sim}
    \vspace{-1.0em}
\end{figure*}

To evaluate the agility and robustness of the proposed method, we conduct extensive simulated and real-world experiments, focusing on maximum flight speed, average speed, and success rate. A trial is considered successful if the UAV flies toward the target direction, exits the obstacle region, and completes the task without any collision.
\subsection{Training Performance}
We train the policy using a CUDA-based differentiable simulator. On a workstation equipped with an Intel i9 CPU and an NVIDIA RTX 4090 GPU, policy convergence is achieved within approximately two hours. The two proposed components---removal of rotational optical flow and incorporation of flow uncertainty mask---significantly improve training performance. As shown in Fig.~\ref{fig:training}, compared with raw optical flow, using translational optical flow results in lower training loss, higher success rates, and improved maximum and average flight speeds. Augmenting translational optical flow with a flow uncertainty mask further enhances performance across all metrics, with a particularly notable gain in success rate.

\subsection{Simulation Experiment}
\subsubsection{AirSim Simulation}
We evaluate the proposed approach in the standard AirSim simulator~\cite{airsim2017fsr}. We consider two representative scenarios: (1) a dense, structured obstacle field consisting of cubes, horizontal beams, and spherical obstacles, designed to evaluate both horizontal and vertical avoidance; and (2) a large open-field environment populated with cylindrical obstacles, intended to assess performance in large-scale scenes. As shown in Fig.~\ref{fig:airsim_sim}(a) and (b), the learned policy demonstrates robust and agile flight in both environments.

\begin{figure}
    \centering
    \includegraphics[width=0.42\textwidth]{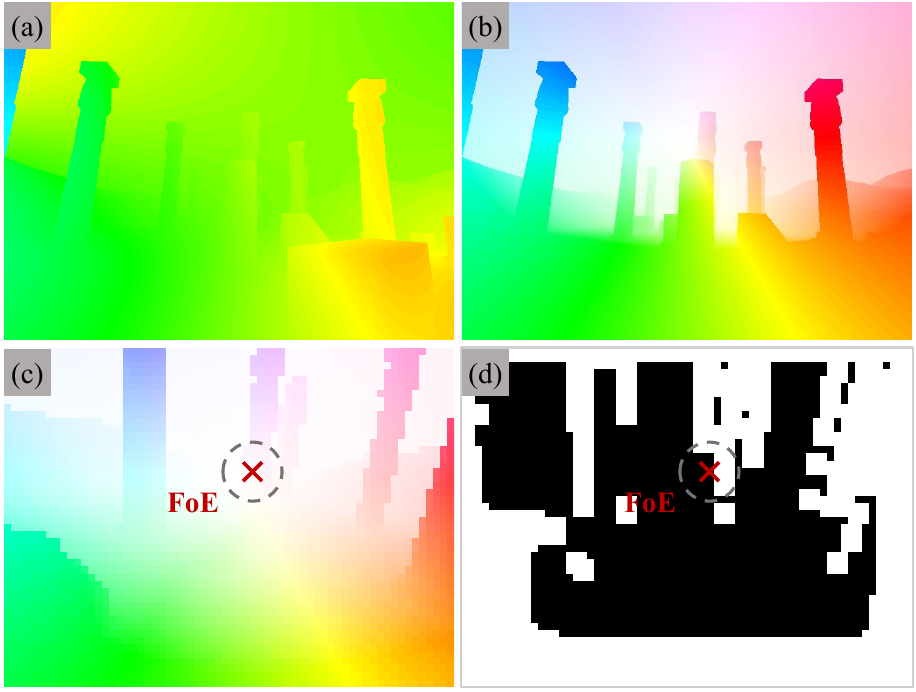}
    \caption{\textbf{Visualization of visual inputs in AirSim simulations:}  (a) Raw optical flow and (b) translational optical flow for a frame dominated by rotational motion during an obstacle avoidance experiment using raw optical flow as input. (c) Translational optical flow and (d) the corresponding uncertainty mask for a representative frame from a trajectory with a target speed of 12~m/s, where the policy input consists of translational optical flow augmented with an uncertainty mask.}
    \label{fig:airsim_vis}
    \vspace{-1.0em}
\end{figure}

To quantify the contribution of each module, we perform ablation studies using three types of visual inputs: raw optical flow, translational optical flow, and translational optical flow augmented with a flow uncertainty mask. For each desired speed, 20 independent trials are conducted, and the results are summarized in Fig.~\ref{fig:airsim_sim}.

In the Structured scenario (Fig.~\ref{fig:airsim_sim}(c) and (d)), policies using translational optical flow consistently outperform those based on raw optical flow in both success rate and maximum speed. By removing rotational components, translational optical flow provides a more consistent motion representation for obstacles and background, improving policy stability at high speed. Incorporating the uncertainty mask further yields the best overall performance across all metrics, consistent with the training results.

In the Open-field scenario (Fig.~\ref{fig:airsim_sim}(c) and (e)), the raw-flow policy struggles to separate obstacles from the background because strong rotations make obstacle-induced flow ambiguous with background motion, as shown in Fig.~\ref{fig:airsim_vis}(a). In contrast, translational optical flow produces clearer obstacle boundaries and better background separation (Fig.~\ref{fig:airsim_vis}(b)), leading to higher success rates, although the raw-flow policy attains slightly higher average and maximum speeds by failing to decelerate near obstacles. At target speeds above 10~m/s, it yields no successful trials. Using translational optical flow alone further improves the success rate, but still fails at 12~m/s because the flow magnitude becomes weak near the FoE (Fig.~\ref{fig:airsim_vis}(c)), causing repeated collisions with the central pillar. Augmenting translational optical flow with the uncertainty mask introduces additional structural cues near the FoE (Fig.~\ref{fig:airsim_vis}(d)), compensates for unreliable flow at high speed, and achieves the best overall performance in both success rate and flight speed.

\subsubsection{Flightmare Simulation}

\begin{figure}
    \centering
    \includegraphics[width=0.45\textwidth]{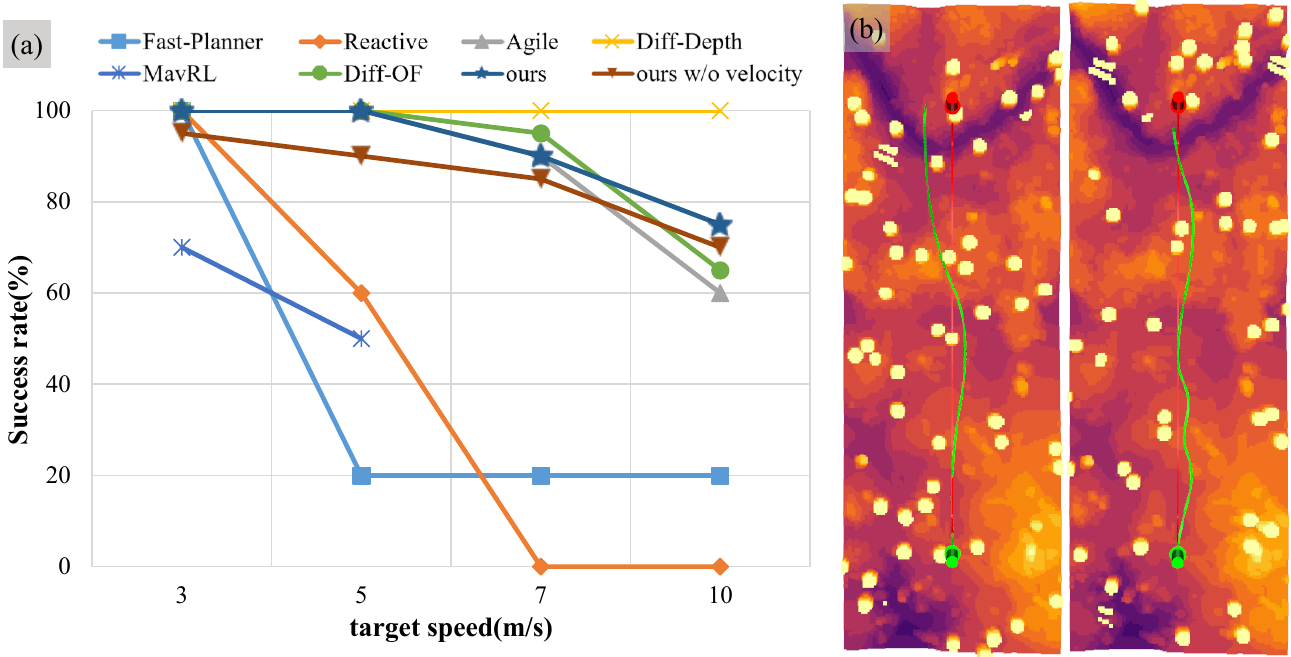}
    \caption{\textbf{Baseline comparison for sim-to-sim transfer:} Each method is evaluated over 20 trials. (a) Success rates of our approach and state-of-the-art baselines under different target speeds in the same simulator. (b) Top-down visualization of representative UAV trajectories in the Flightmare test environment. The background colormap indicates the environment height map.}
    \label{fig:flightmare}
    \vspace{-1.0em}
\end{figure}

To evaluate generalization, we deploy the trained policy in a previously unseen forest environment in Flightmare~\cite{song2020flightmare}. Representative top-down trajectories are shown in Fig.~\ref{fig:flightmare}, illustrating effective obstacle avoidance under varying obstacle densities. We further compare our method with the depth-based Agile~\cite{loquercio_learning_2021}, the mapping-based Fast-Planner~\cite{zhou2019fast}, the reactive planner Reactive~\cite{florence2020integrated}, the optical-flow-based Diff-OF~\cite{huyu2025SeingThroughPixelMotion}, the depth-based differentiable-learning method Diff-Depth~\cite{zhang2025learning}, and the reinforcement-learning-based adaptive-speed method MAVRL~\cite{Yu2025MAVRL}. As shown in Fig.~\ref{fig:flightmare}(a), our method consistently outperforms Fast-Planner, Reactive, and MAVRL, while remaining competitive with Agile, Diff-OF, and Diff-Depth. In particular, although Agile and Diff-OF perform competitively at lower target speeds, our method shows stronger robustness in the high-speed regime.

\subsection{Real-World Experiment}
\begin{figure*}
    \centering
    \includegraphics[width=0.82\textwidth]{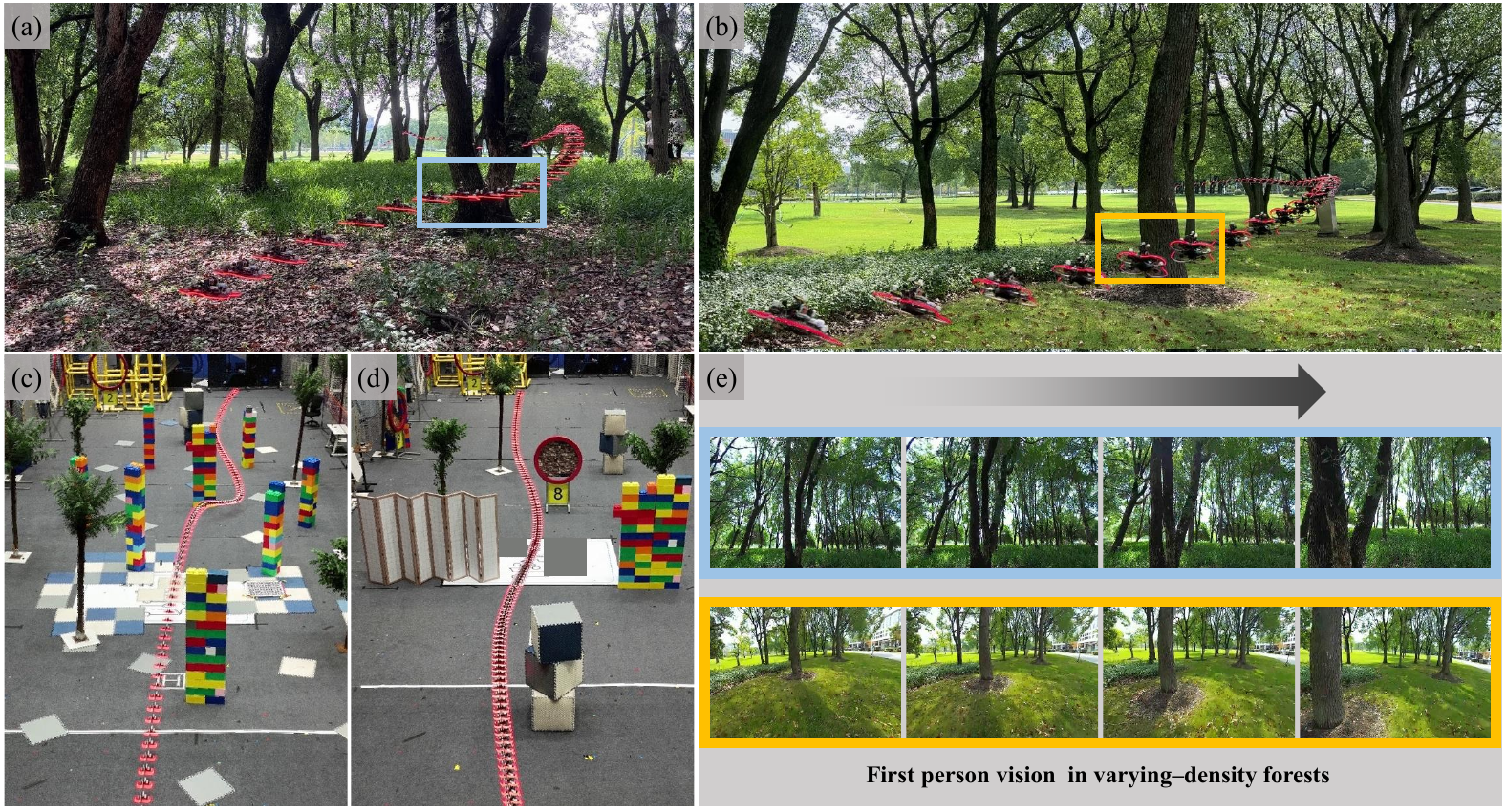}
    \caption{\textbf{Real-world experimental results:} We evaluate our method in previously unseen real environments, including (a) a dense forest, (b) a sparse forest, and (c)-(d) cluttered indoor scenes. (e) illustrates representative trunk-avoidance behaviors. Notably, in the sparse forest environment (b), the quadrotor achieves a peak speed of 11.79~m/s. UAV trajectories are visualized by overlaying the estimated positions on the video frames (see the supplementary video for details).}
    \label{fig:real_exp}
    \vspace{-1.0em}
\end{figure*}
To improve practical deployability, we introduce a no-external-velocity mode that removes the explicit velocity state from the policy input, while retaining standard IMU-based attitude estimation and low-level stabilization on the flight controller. Instead of directly estimating velocity from optical flow, the network \emph{implicitly infers} translational velocity from visual inputs, control outputs, and the UAV dynamics model. During training, an auxiliary supervision term enforces consistency between the inferred and ground-truth motion under the system dynamics. This enables the policy to learn a visually grounded ego-motion representation, improving robustness when onboard state estimation is noisy or unreliable. We further evaluate this no-external-velocity variant in simulation. The results show only a modest performance drop relative to the full-input policy, supporting its practicality for real-world deployment, as shown in Fig.~\ref{fig:flightmare}.

The trained policy is deployed on a real-world offboard platform. FPV images are transmitted wirelessly to a local host (a laptop with an RTX 5090 GPU), where optical-flow estimation and policy inference are performed, while control commands and IMU measurements are exchanged with the UAV flight controller over the same wireless link. In practice, optical-flow estimation, rather than policy inference, is the main computational bottleneck, motivating the offboard setup in real-world experiments. Experiments are conducted in both indoor and outdoor environments. The indoor setup includes pillars, boxes, screens, and artificial trees to emulate cluttered scenes, whereas outdoor tests are performed in a natural forest. As shown in Fig.~\ref{fig:real_exp}, the policy achieves agile obstacle avoidance in both settings, reaching maximum speeds of 8.11~m/s indoors and 11.79~m/s outdoors. Outdoor trials attain a success rate of 93.3\% over 30 real-world flight trials; the failure cases are caused by a large optical-flow estimation error under a sudden illumination change during traversal. To the best of our knowledge, this is the first monocular-vision-based UAV flight exceeding 10~m/s in real-world environments. Notably, the deployed policy requires no fine-tuning and is identical to that used in simulation.

For real-world deployment, we evaluate several state-of-the-art optical-flow networks, including GMFlow~\cite{xu2022gmflow}, SeaRaft~\cite{wang2024sea}, and FlowFormer~\cite{huang2023flowformer}, and select GMFlow for its practical balance between estimation quality and runtime efficiency in our real-time offboard pipeline. Nevertheless, the estimated flows still differ notably from the simulation ground truth, particularly in the translational component. Moreover, because the uncertainty mask depends on both forward and backward flows, these errors can be further amplified. These results indicate that accurate real-time optical flow remains essential for robust real-world deployment.
\section{Conclusion}

Our method removes the rotational component from the optical flow and uses the translational flow for input.
This enhances the network’s understanding of the scene and improves UAV agility. The introduction of the flow uncertainty mask mechanism provides additional structural information about the environment, particularly in the FoE region, further enhancing UAV agility and significantly improving obstacle avoidance robustness. Simulation and ablation experiments validate the generalization ability of the method and the effectiveness of its individual modules.
Real-world experiments demonstrate that our method maintains strong agile obstacle avoidance capabilities.

\textbf{Limitations:} While we improve autonomous FPV flight performance, experiments are still conducted on an offboard system due to the high computational demands of optical flow networks. This reliance on external processing limits real-time applicability onboard. Moreover, both structural cues depend on flow, so the quality of estimated optical flow critically affects FPV agility and robustness. Future work should enhance obstacle flow accuracy and reduce optical flow network complexity.

\ifCLASSOPTIONcaptionsoff
  \newpage
\fi

\balance
\bibliographystyle{ieeetr}
\bibliography{reference}

\end{document}